\documentclass{bmvc2k}
\usepackage{booktabs}

\title{Better Retrieval, Not Better Binding: Pretraining Exposure in Egocentric Video Retrieval}

\addauthor{Yuqi Guo}{rr24582@bristol.ac.uk}{1}

\addauthor{Michael Wray}{michael.wray@bristol.ac.uk}{1}

\addinstitution{
 University of Bristol\\
 Bristol, UK
}

\runninghead{}{Better Retrieval, Not Better Binding}

\makeatletter
\def\bmv@RenderAuthInstTwoColumn{%
  \ifnum\bmv@ninsts>1
    \def\bmv@FN{FN}%
  \else
    \def\bmv@FN{}%
  \fi
  \noindent\hspace*{24pt}%
  \begin{minipage}[t]{\dimexpr0.49\textwidth-24pt\relax}%
    \def\bmv@action##1{%
      \bmv@unhbox{authname\bmv@FN##1}\\%
      \bmv@unhbox{authmail##1}%
    }%
    \def\bmv@between{\\[3pt]}%
    \expandafter\bmv@maplistaux\bmv@auths\bmv@endstop
  \end{minipage}%
  \bmv@RAIspace\hfill
  \begin{minipage}[t]{0.48\textwidth}%
    \def\bmv@action##1{%
      \bmv@unhbox{inst\bmv@FN##1}%
    }%
    \def\bmv@between{\\[4pt]}%
    \expandafter\bmv@maplistaux\bmv@insts\bmv@endstop
  \end{minipage}%
}
\makeatother

\begin{document}

\maketitle

\begin{abstract}
Video-language models (VLMs) achieve strong performance on video--text retrieval and are increasingly used for egocentric video understanding. Yet, despite gains in zero-shot retrieval, they have shown limitations in compositional understanding, and it remains unclear whether they understand compositional relationships between verbs and nouns.

In this work, we introduce Ego-ExBind, a compositional diagnostic protocol for analyzing pretraining exposure and verb--noun binding in egocentric video--text retrieval. Ego-ExBind consists of an Exposure Ledger built from the EgoClip pretraining corpus used by EgoVLPv2, which defines Seen Composition, Unseen Composition, and Unseen Atom splits for EPIC-KITCHENS-100 multi-instance retrieval (MIR). We further probe noun and verb binding channels through controlled noun-swap and verb-swap negatives.

We find that zero-shot retrieval is positively associated with pair
exposure within seen compositions, even after controlling for the
marginal frequencies of verbs and nouns. However, higher pair exposure
is associated with larger noun-swap margins and slightly smaller
verb-swap margins. This asymmetry motivates us to test whether exposure
reweighting can improve retrieval with little change in the measured
swap margins.

To isolate the effect of exposure input, we keep the trained checkpoint fixed and replace only the exposure input at inference. We demonstrate that this exposure reweighting can improve video--text retrieval without stronger verb--noun compositional understanding, and that the gain depends on the correct sample-level exposure correspondence. By separating exposure-driven retrieval gains from verb--noun binding, Ego-ExBind provides a diagnostic framework for evaluating whether future improvements in egocentric MIR reflect compositional understanding rather than pretraining familiarity.

\end{abstract}

%-------------------------------------------------------------------------
\section{Introduction}
\label{sec:introduction}

Vision-language models (VLMs) achieve impressive performance on
video--text retrieval tasks and transfer effectively to downstream
egocentric settings~\cite{lin2022egocentric,zhao2023learning,pramanick2023egovlpv2}.
Yet, it remains unclear whether high retrieval
scores on these downstream benchmarks indicate that the model can generalize
to verb--noun compositions, or whether this performance is associated with
exposure to the same compositions in the pretraining data. Zero-shot
evaluation rules out downstream adaptation, but it does not test
pretraining exposure to the verbs, nouns, and compositions being evaluated.

Recent literature has therefore evaluated the compositional ability of VLMs
using controlled retrieval and matching
tasks~\cite{thrush2022winoground,yuksekgonul2023when}.
CREPE further studies generalization across seen and unseen
compositions~\cite{ma2023crepe}, while recent work shows that pretraining concept frequency can predict
downstream performance~\cite{udandarao2024no}.
However, these analyses either probe whether models understand compositional
structure or characterize how performance varies with pretraining exposure.
They do not establish whether the retrieval benefit associated with
composition-level exposure corresponds to how well models match actions to their corresponding objects.

This ambiguity is particularly important for egocentric retrieval because standard retrieval evaluation does not explicitly require correct verb--noun binding.
In EK100-MIR, relevance is computed from verb and noun overlap
separately~\cite{damen2022rescaling}. 
At the same time, recent work on egocentric
video--text retrieval has largely focused on improving retrieval performance,
leading to increasingly competitive mAP and nDCG
results~\cite{wang2024symmetric,he2025contextrefine,li2026tempret}.
Thus, high retrieval scores may not reflect correct verb--noun binding.

In this work, we introduce Ego-ExBind, a diagnostic protocol for analyzing
pretraining exposure and verb--noun binding in egocentric video--text
retrieval. We introduce an Exposure Ledger to characterize what the model has
seen during pretraining, and a Binding Probe based on controlled noun-swap and
verb-swap negatives to characterize what compositional structure the model has
learned. Using these evaluations on a frozen EgoVLPv2 model, we find that
retrieval performance is consistently associated with verb-noun pair exposure and is
higher for Seen Composition queries than for Unseen Composition and Unseen
Atom queries. However, this retrieval advantage is not accompanied by a corresponding improvement in binding: noun binding is associated with exposure, whereas verb
binding shows no corresponding positive association.

Standard downstream
adaptation and targeted binding interventions can improve retrieval-oriented
metrics, yet fail to consistently repair verb--noun binding. More importantly, with the trained checkpoint fixed during counterfactual evaluation, exposure-dependent score reweighting can improve retrieval while leaving binding largely unchanged. Counterfactual tests further show that this retrieval gain depends on the correct sample exposure correspondence.
Thus,
improved retrieval performance should not, by itself, be interpreted as
evidence of stronger compositional generalization.

Our main contributions are threefold:
\begin{enumerate}
    \item \textbf{A diagnostic protocol for jointly measuring pretraining
    exposure and verb--noun binding.}
    An Exposure Ledger characterizes what the model has seen, while controlled
    noun-swap and verb-swap probes characterize what compositional structure it
    has learned.
    \item \textbf{A diagnosis of the mismatch between exposure-dependent
    retrieval and verb--noun binding.}
    Retrieval performance is consistently associated with pair exposure,
    whereas noun-binding and verb-binding margins behave asymmetrically, and
    retrieval-oriented adaptation fails to consistently repair verb--noun
    binding.
    \item \textbf{Controlled evidence that retrieval gains can arise from
    exposure reweighting without compositional repair.}
    With the trained checkpoint fixed during counterfactual evaluation, exposure reweighting and counterfactual tests show that retrieval can benefit from the true exposure signal without a corresponding improvement in verb--noun binding.
\end{enumerate}
\section{Related Work}
\label{sec:related-work}

\paragraph{Compositional evaluation of vision-language models.}
Vision-language model evaluation has attracted increasing attention, and
recent work evaluates compositional understanding through controlled changes
in lexical content, relations, attributes, order, and
representations~\cite{thrush2022winoground,yuksekgonul2023when,cascante2023going,kamath2024hard}. Kamath et al.~\cite{kamath2024hard} further show that compositional
information such as object relationships and attribute-object associations
can be lost in pooled text representations, and that this loss correlates
with multimodal matching errors.
Our compositional evaluation is most closely related to
CREPE~\cite{ma2023crepe}. CREPE evaluates systematic generalization
across seen and unseen compositions. In egocentric video retrieval,
we examine how exposure to verb--noun pairs relates to retrieval
performance and to the model's preferences under noun and verb swaps.

Our probe also differs from Winoground~\cite{thrush2022winoground} by dissociating
noun binding from verb binding through separate noun-swap and verb-swap
negatives, allowing us to ask whether exposure-related retrieval advantages
correspond to stronger binding in egocentric video retrieval.

\paragraph{Concept frequency and shortcut learning.}
There are a number of works highlighting shortcut learning across deep
learning, medical AI, and NLP, where models can achieve strong performance by
relying on non-target cues or
heuristics~\cite{geirhos2020shortcut,ong2024shortcut,mccoy2019right}.
Similar shortcut behavior has been observed in vision-language models, where
models can exploit language distributions without image evidence or fail to
preserve compositional binding
in cross-modal alignment~\cite{lin2023revisiting,koishigarina2026clip}.
Most closely related to our exposure analysis, Udandarao et
al.~\cite{udandarao2024no} show a strong relationship between pretraining
concept frequency and downstream multimodal performance.
Although they consider various concept types, their analysis establishes an
overall concept frequency--performance relationship; in contrast, we measure
verb, noun, and verb--noun pair exposure separately and ask whether
exposure-related retrieval gains reflect stronger compositional binding,
rather than merely higher performance.

\paragraph{Composition-aware interventions and counterfactual diagnostics.}
Composition-aware hard negatives have been used to improve compositional or
verb understanding in vision-language
models~\cite{yuksekgonul2023when,momeni2023verbs}, while
invariance-based objectives provide related tools for reducing reliance on
specific signals~\cite{arjovsky2019invariant}. We additionally use lightweight
adaptation as a controlled diagnostic intervention.
Whereas VFC~\cite{momeni2023verbs} uses verb hard negatives to improve verb understanding, we use
the same construction diagnostically to test whether verb binding can
actually be improved.
Kamath et al.~\cite{kamath2024hard} show that gains from
hard-negative fine-tuning can be overstated and can induce oversensitivity to
semantic-preserving perturbations, while Hsieh et
al.~\cite{hsieh2023sugarcrepe} show that hard-negative benchmarks can contain
exploitable biases.
Unlike these works, we use interventions diagnostically. We include an
exposure decorrelation penalty between the noun margin and pair exposure and,
inspired by counterfactual probing such as amnesic
probing~\cite{elazar2021amnesic}, manipulate the exposure signal to test
whether retrieval gains depend on exposure without a corresponding
improvement in binding.

\section{Method}
\label{sec:method}

Most VLMs are consistently evaluated on video-to-text and
text-to-video retrieval and achieve impressive zero-shot performance. We quantify exposure as the number of times an action concept appears in the EgoClip pretraining corpus~\cite{lin2022egocentric}. For a verb $v$, a noun $n$, and a verb--noun pair $(v,n)$, we denote their exposure by frequencies $f_v(v)$, $f_n(n)$, and $f_p(v,n)$, respectively. $f_v$ and $f_n$ count atoms marginally, whereas $f_p(v,n)$ counts \emph{valid} action compositions. In this section, we demonstrate why retrieval score alone is incomplete: retrieval on
EK100~\cite{damen2022rescaling} may reward exposure to individual action concepts or frequently
observed verb--noun pairs rather than their correct composition.

\begin{figure*}[t]
    \centering
    \includegraphics[width=0.98\textwidth]{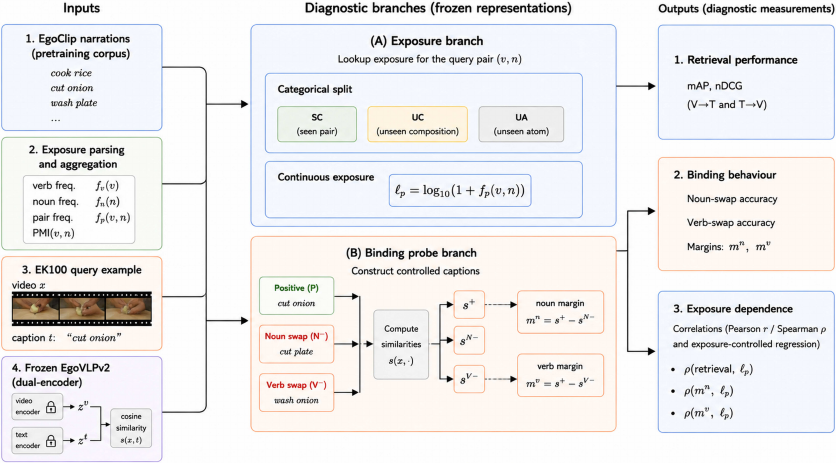}
    \caption{Overview of the Ego-ExBind diagnostic protocol.
    The exposure branch characterises pretraining exposure using an
    Exposure Ledger, whereas the binding-probe branch evaluates noun
    and verb binding using controlled caption swaps.}
    \label{fig:egoexbind-overview}
\end{figure*}

First, we establish verb, noun, and verb--noun pair exposure over the EgoClip
pretraining corpus and examine their relationships with zero-shot retrieval
and binding performance (Section~\ref{sec:exposure-ledger}). We introduce a sequence of controlled
interventions to test whether standard retrieval adaptation can induce these
expected behaviors (Section~\ref{sec:diagnostic-chain}). Finally, we manipulate
exposure directly at the scoring stage to determine whether retrieval gains
depend on exposure-conditioned score reweighting.
Figure~\ref{fig:egoexbind-overview} summarizes the diagnostic protocol of
Ego-ExBind. Given a frozen EgoVLPv2 backbone~\cite{pramanick2023egovlpv2}, we apply a two-branch diagnostic protocol
under the zero-shot dual-encoder setting and use the video--text similarity
for evaluation. The two branches are designed to distinguish the effect of
pretraining exposure from genuine verb--noun compositional binding in strong
retrieval performance. This distinction cannot be determined from standard
retrieval metrics alone, because they do not explicitly measure whether the
correct verb and noun are jointly bound.

\subsection{Exposure Ledger and Diagnostic Protocol}
\label{sec:exposure-ledger}

To evaluate compositional generalization in egocentric retrieval, we adopt a
minimal compositional language over actions. Accordingly, an atom corresponds
to a single action concept and is subtyped into two types: a verb $v$, such as
\emph{cut}, and a noun $n$, such as \emph{onion}. An action composition is the
binding of a verb to a noun, written as $c=(v,n)$, such as \emph{cut onion}.

This restriction allows verb recognition, noun recognition, and the binding
between them to be measured in isolation.

\paragraph{Retrieval relevance and its limitation.}
Standard evaluations report strong zero-shot retrieval scores for EgoVLPv2 on
EK100; however, these retrieval metrics do not necessarily measure
compositional understanding. Here, zero-shot means that the pretrained
backbone is directly applied to video--text retrieval without fine-tuning on
the downstream EK100 dataset.

EK100 multi-instance retrieval evaluates predictions using an official graded
relevance matrix~\cite{damen2022rescaling}. Let $X_i^V$ and $X_i^N$ denote the sets of verb and noun
classes associated with video $x_i$, and let $C_j^V$ and $C_j^N$ denote those
associated with caption $c_j$. The official relevance is defined as
\begin{equation}
\mathcal{R}(x_i,c_j)
=
\frac{1}{2}
\left(
\frac{\left|X_i^V\cap C_j^V\right|}
{\left|X_i^V\cup C_j^V\right|}
+
\frac{\left|X_i^N\cap C_j^N\right|}
{\left|X_i^N\cup C_j^N\right|}
\right).
\label{eq:official-relevance}
\end{equation}

The relevance is therefore the mean Intersection-over-Union of the verb and
noun classes. Although this formulation supports graded semantic relevance,
the verb and noun overlaps are evaluated independently. It does not contain an
explicit term requiring the correct verb--noun pair to be matched.
Consequently, a candidate containing the correct noun but the wrong verb, or
the correct verb but the wrong noun, may still receive partial relevance.
High retrieval performance alone therefore does not establish correct
verb--noun binding.

\paragraph{Exposure statistics and exposure-relative splits.}
EgoClip is the egocentric video--text corpus used to pretrain EgoVLPv2~\cite{pramanick2023egovlpv2}. We therefore use EgoClip for measuring the model's pretraining exposure to EK100 verbs, nouns, and verb--noun compositions. Even though EgoClip and EK100 both contain egocentric kitchen scenes, an EK100
action may have different levels of exposure in the EgoClip pretraining
corpus. To systematically measure this exposure, for each action composition
$c=(v,n)$, we count the pair frequency $f_p(v,n)$, the verb frequency $f_v(v)$,
and the noun frequency $f_n(n)$. Based on these exposure statistics, we define three mutually exclusive evaluation splits: 
\begin{equation}
\begin{aligned}
\mathcal{D}_{\mathrm{SC}}
&=
\{(x_i,t_i):f_p(v_i,n_i)>0\},\\
\mathcal{D}_{\mathrm{UC}}
&=
\{(x_i,t_i):
f_p(v_i,n_i)=0,\,
f_v(v_i)>0,\,
f_n(n_i)>0\},\\
\mathcal{D}_{\mathrm{UA}}
&=
\{(x_i,t_i):
f_v(v_i)=0
\ \lor\
f_n(n_i)=0\}.
\end{aligned}
\label{eq:exposure-splits}
\end{equation}

Seen Composition (SC) contains verb--noun pairs that have been observed in the
EgoClip pretraining corpus. Unseen Composition (UC) contains pairs that have
not been observed together, although both constituent atoms have been
observed individually. UC therefore evaluates whether the model can recombine
familiar verbs and nouns into unseen action compositions~\cite{hupkes2020compositionality}. Unseen Atom (UA)
contains actions for which either the verb or the noun has not been observed
during pretraining.

Beyond raw frequency, we quantify the strength of each verb--noun association
using pointwise mutual information:
\begin{equation}
\operatorname{PMI}(v,n)
=
\log
\frac{\hat{P}(v,n)}
{\hat{P}(v)\hat{P}(n)}.
\label{eq:pmi}
\end{equation}

PMI measures how strongly a verb and a noun are associated by comparing
their observed co-occurrence probability with the probability expected
under independence, given their marginal frequencies. We compute PMI
only for compositions observed in the Exposure Ledger, for which
$\hat{P}(v,n)>0$. As in the continuous exposure analysis, we use
verb--noun pairs as the units of analysis and include only Seen
Composition (SC) pairs with at least three evaluation queries.

In all cases, ``seen'' and ``unseen'' are defined relative to the EgoClip
pretraining corpus rather than downstream EK100 adaptation.

\paragraph{Frozen exposure ledger construction.}
To parse EgoClip into verbs, nouns, and verb--noun pairs, we extract
action--object candidates from the narrations using spaCy and map them into the
EK100 action taxonomy containing 97 verb classes and 300 noun classes~\cite{damen2022rescaling}.

We count an action as pair-level evidence only when both the verb and noun are
confidently mapped and form a valid action--object pair. If only one atom can
be confidently mapped, it contributes only to the corresponding marginal
frequency and is not used as pair-level evidence. This conservative rule
prevents parser fallback from creating false verb--noun compositions.

We additionally perform a manual audit of high-risk mappings produced by
parser fallback and taxonomy normalization. After construction and auditing,
the exposure ledger is frozen and reused unchanged across all subsequent
retrieval, binding, and intervention experiments. The resulting EK100
evaluation set contains 9,668 queries, including 8,648 SC, 713 UC, and 307 UA
queries.

\paragraph{Exposure-stratified and continuous retrieval analysis.}
We measure zero-shot retrieval exposure dependence in two ways. First, we
compare official V$\rightarrow$T and T$\rightarrow$V mAP and nDCG across SC,
UC, and UA. If the model can systematically recombine familiar action atoms,
the performance difference between SC and UC should remain limited, because
both splits contain previously observed verbs and nouns.

Second, SC, UC, and UA splits reveal whether a composition has been observed in pretraining, rather than the amount of exposure. Therefore, our goal is to examine whether verb--noun pair frequency predicts retrieval performance beyond the marginal frequencies of the verb and noun. We therefore examine continuous exposure dependence. For each pair-level retrieval metric $y_i$, such as mAP or nDCG, we fit
\begin{equation}
\begin{aligned}
y_i
={}&
\alpha
+
\beta_p\log_{10}\!\left(1+f_p(v_i,n_i)\right)\\
&+
\beta_v\log_{10}\!\left(1+f_v(v_i)\right)
+
\beta_n\log_{10}\!\left(1+f_n(n_i)\right)
+
\epsilon_i.
\end{aligned}
\label{eq:exposure-regression}
\end{equation}

Here, $i$ indexes an eligible verb--noun pair, $y_i$ denotes its pair-level retrieval score, 
$\alpha$ is the intercept, and $\beta_p$, $\beta_v$, and $\beta_n$ are the 
coefficients for pair, verb, and noun exposure, respectively. 
The residual $\epsilon_i$ captures variation not explained by these exposure variables.

We report Pearson correlation, Spearman rank correlation, and the OLS
coefficient for pair exposure. Controlling for verb and noun marginal
frequencies allows us to determine whether pair exposure predicts retrieval
performance beyond the popularity of the individual action atoms.

\paragraph{Verb--noun binding probes.}
Given the weaker discriminative capability of the verb channel, it is of
interest to determine whether verb understanding and binding can be
strengthened using targeted hard negatives. Recent work on verb-focused hard-negative training~\cite{momeni2023verbs} provides good inspiration. We target the verb channel by
keeping the noun fixed and replacing only the verb, allowing verb discrimination
to be evaluated independently of the noun channel.

Building upon the exposure-relative evaluation, we leverage binding channel
dissociation and generate noun-swap and verb-swap probes. Given a video and its
positive action description $t_i^+=(v_i,n_i)$, we construct two controlled
alternatives.

For the noun-swap condition, we keep the verb unchanged and replace the noun,
for example, \emph{cut onion} versus \emph{cut tomato}. For the verb-swap
condition, we keep the noun unchanged and replace the verb, for example,
\emph{cut onion} versus \emph{wash onion}. Each test case therefore contains
one video and three grammatically matched texts: the correct action text, a
noun-swapped text, and a verb-swapped text.

Given the model similarity $s(x,t)$, we define the noun and verb margins as
\begin{equation}
m_i^n
=
s(x_i,t_i^+)-s(x_i,t_i^{n-}),
\qquad
m_i^v
=
s(x_i,t_i^+)-s(x_i,t_i^{v-}).
\label{eq:binding-margins}
\end{equation}

A positive noun margin indicates that the model prefers the correct noun while
the verb is fixed. A positive verb margin indicates that the model prefers the
correct verb while the noun is fixed.

We use two complementary probe scales. The small controlled probe contains
200 query videos and three encoded texts per query, resulting in 600 texts
in total. It is used to report discrete noun and verb accuracies and their
corresponding margins across the exposure-relative groups.

The large binding probe applies the same noun-swap and verb-swap construction
to all 8,592 eligible SC queries in the EK100 validation set. It is used for
continuous margin exposure analysis, providing substantially greater
statistical power than the small controlled probe. For both probes, we examine
the relationships between noun and verb margins and log-scaled pair exposure.

\subsection{Exposure-Driven Retrieval: A Diagnostic Chain}
\label{sec:diagnostic-chain}

We hypothesize that the failure to preserve binding is
attributable to the retrieval objective itself, rather than to insufficient
adaptation.

\begin{figure*}[t]
    \centering
    \includegraphics[width=0.6\textwidth]
    {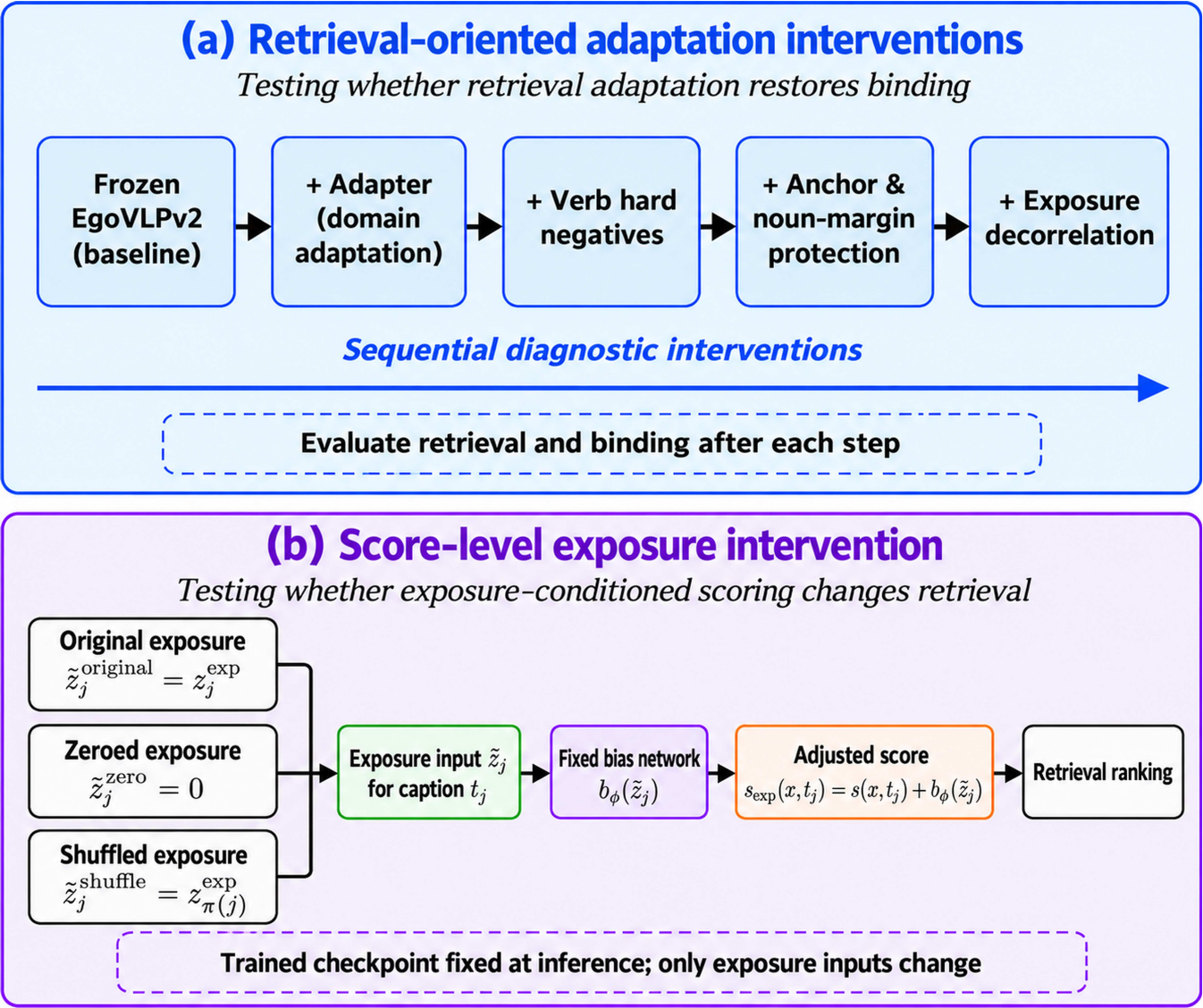}
    \caption{Two families of controlled interventions used in Ego-ExBind.
    (a) Retrieval-oriented adaptation interventions test whether retrieval
    adaptation restores verb--noun binding.
    (b) The score-level intervention tests whether exposure can alter retrieval with the trained checkpoint fixed.}
    \label{fig:diagnostic-interventions}
\end{figure*}

To understand this relationship, we introduce a controlled diagnostic chain
between domain adaptation, targeted verb discrimination, margin preservation,
and exposure dependence. For each intervention, we evaluate both official
retrieval metrics and the noun/verb binding probes. Therefore, an improvement
in retrieval is not treated as evidence of repaired binding unless the
corresponding binding behavior also improves.
Figure~\ref{fig:diagnostic-interventions} summarises the two families of
controlled interventions used in Ego-ExBind.

\paragraph{Frozen zero-shot baseline.}
We first evaluate the pretrained EgoVLPv2 model directly on EK100 without
updating any model parameters. This frozen model provides the reference
retrieval scores and noun/verb binding margins for all subsequent
interventions.

\paragraph{Domain adaptation control.}
Compared with full fine-tuning, domain adapters provide a lightweight form of
adaptation with fewer trainable parameters while keeping
the pretrained backbone frozen, making them more suitable for diagnosing
whether retrieval improvement corresponds to binding repair. The zero-initialized
residual adapters begin from the same similarities as the frozen baseline,
making the two directly comparable, while allowing adaptation to the EK100
retrieval task.

Starting from the frozen EgoVLPv2 backbone, we add residual
adapters~\cite{gao2024clip} to the video and text representations.
We train only the adapters using the official EK100
soft-relevance objective derived from $\mathcal{R}$. This intervention tests
whether adaptation to the downstream retrieval objective improves
compositional binding or mainly calibrates the model toward the EK100
relevance structure.

\paragraph{Verb-targeted hard negatives.}
Inspired by composition-aware and verb-focused hard-negative
training~\cite{yuksekgonul2023when,momeni2023verbs}, for each positive
video--caption pair, we generate a negative caption by preserving the noun
and replacing only the verb.

The generated negative is added only
to the denominator of the corresponding V$\rightarrow$T anchor, while the
original soft relevance loss is retained. Because the noun is fixed and only
the verb is changed, this intervention specifically tests whether targeted
retrieval pressure can strengthen the weak verb-discrimination channel.

\paragraph{Similarity anchoring and noun-margin protection.}
A targeted intervention may be difficult to interpret if it simultaneously
changes the entire similarity space or degrades the existing noun channel. We
therefore start from the domain-adapted model and add a similarity anchor term
that keeps the current similarity matrix close to the adapted reference.

We additionally introduce a noun-margin protection term that prevents the
mean noun margin from falling below a predefined floor.

\paragraph{Exposure decorrelation.}
We apply exposure decorrelation to the noun margin rather than to the verb margin. Motivated by invariance-based learning~\cite{arjovsky2019invariant}, we add a
decorrelation penalty that suppresses the Pearson
correlation across samples in each batch between the noun margin and log-scaled pair exposure. This intervention does
not directly attempt to repair verb binding. Instead, it asks whether
separating the noun-binding channel from pair
exposure, while preserving its discriminative margin, leads to more
compositionally faithful binding.

The complete objective is
\begin{equation}
\mathcal{L}
=
\mathcal{L}_{\mathrm{base}}
+
\lambda_a \left\| S - S_{\mathrm{DA}} \right\|_F^2
+
\lambda_p \max(0, \delta_n - \mathrm{mean}(\mathbf{m}^n))
+
\lambda_d \rho(\mathbf{m}^n, \boldsymbol{\ell}_p)^2 ,
\end{equation}
where
\begin{equation}
\boldsymbol{\ell}_p =
\left[
\log_{10}(1+f_p(v_i,n_i))
\right]_{i=1}^{B}.
\end{equation}

Here, $S_{\mathrm{DA}}$ denotes the similarity matrix produced by the
domain-adapted reference model, $\delta_n$ is the noun-margin protection floor,
$\lambda_a$, $\lambda_p$, and $\lambda_d$ are the weights for the similarity
anchoring, noun-margin protection, and exposure decorrelation terms, respectively,
$\boldsymbol{\ell}_p$ is the vector of log-scaled pair exposure within the batch,
and $\rho$ is the Pearson correlation. We evaluate different values of
$\lambda_d$ to determine whether reducing exposure dependence also changes the
underlying binding behavior.

Overall, these interventions form a diagnostic chain rather than a collection
of independent model contributions. They test whether ordinary domain
adaptation, targeted verb discrimination, noun-margin preservation, or noun
exposure decorrelation is sufficient to make retrieval improvements
correspond to improved compositional binding.

\subsection{Exposure Manipulation and Counterfactual Verification}
\label{sec:exposure-manipulation}

To understand what it takes for exposure to affect retrieval and binding, we
next manipulate exposure directly. While prior work establishes a
correlational relationship between pretraining concept frequency and
downstream performance~\cite{udandarao2024no}, we evaluate this relationship
at the compositional level. To determine whether retrieval at a fixed trained checkpoint depends on exposure information, we jointly train the video and text residual adapters with the exposure-bias network while keeping the pretrained EgoVLPv2 backbone frozen. We then fix the trained adapters and bias network and manipulate only the exposure input during counterfactual evaluation. First, we evaluate the exposure-conditioned score adjustment, and then replace the exposure input with its original, zeroed, or shuffled variant to determine whether any gain requires the correct sample--exposure pairing.

\paragraph{Exposure score bias.}
We add a centered text-side exposure bias to the similarity scores produced by the jointly trained residual adapters:
\begin{equation}
s_{\mathrm{exp}}(x_i,t_j)
=
s(x_i,t_j)
+
b_{\phi}(\mathbf{z}^{\mathrm{exp}}_j),
\label{eq:exposure-score}
\end{equation}
where $\mathbf{z}^{\mathrm{exp}}_j$ denotes the exposure feature vector associated with candidate text $t_j$, containing the pair frequency, verb frequency, noun frequency, and PMI. 
Here, $s(x_i,t_j)$ denotes the adapted similarity of the jointly trained
checkpoint, and $b_\phi(\mathbf{z}^{\mathrm{exp}}_j)$ denotes the centered
exposure-bias contribution.
The exposure conditioning network
$b_\phi$ is zero-initialized, following the adaLN-Zero initialization
principle~\cite{peebles2023scalable}, so that its bias is initially zero
and leaves the base similarity scores unchanged.

We do not include SC, UC, or UA indicators in the exposure input. Therefore,
the intervention cannot directly use evaluation split labels.

During counterfactual evaluation, the trained adapters and bias network
remain fixed; differences among Original, Zeroed, and Shuffled therefore
isolate the effect of replacing the exposure input within the same
checkpoint.

\paragraph{Counterfactual exposure intervention.}
To evaluate whether the retrieval gain depends on the true sample--exposure
relationship, we replace the exposure input at inference while keeping the
jointly trained adapters and bias network fixed. Let $\pi$ denote a random permutation of candidate indices across the evaluation set.
We define three counterfactual exposure inputs:
\begin{equation}
\tilde{\mathbf{z}}_j^{\mathrm{original}}
=
\mathbf{z}_j^{\mathrm{exp}},
\qquad
\tilde{\mathbf{z}}_j^{\mathrm{zero}}
=
\mathbf{0},
\qquad
\tilde{\mathbf{z}}_j^{\mathrm{shuffle}}
=
\mathbf{z}_{\pi(j)}^{\mathrm{exp}}.
\label{eq:counterfactual-exposure}
\end{equation}

In the original condition, each text retains its true exposure statistics. In
the zeroed condition, the exposure signal is removed by setting all exposure
features to zero before standardisation; all candidates therefore receive the
same standardised exposure input, so the bias cancels after centering. Consequently, Zeroed recovers the adapter-only scores of the same trained checkpoint, rather than the original frozen zero-shot baseline. In the shuffled condition, the empirical exposure distribution is preserved, but the
correspondence between each text and its exposure is broken by assigning it
the exposure statistics of another sample. Shuffling is performed across
candidates in the evaluation set.

We evaluate both retrieval performance and binding margins under all three
conditions. Zeroing tests whether the exposure pathway is necessary for the
learned score adjustment. Shuffling tests whether the adjustment depends on
the correct sample--exposure pairing rather than only on the overall exposure
distribution.

Because the trained checkpoint remains fixed across these counterfactual
conditions, the intervention isolates the effect of exposure-conditioned
score reweighting. The causal interpretation is therefore limited to this
scoring pathway: it tests whether changing the exposure input can alter retrieval
scores when the trained checkpoint is fixed. It does not claim that changing
exposure in the original pretraining corpus would necessarily change the
learned representations, because the backbone is not retrained under
different exposure distributions.
\section{Experiments}
\label{sec:experiments}

\subsection{Experimental Setup}
\label{sec:experimental-setup}

\paragraph{Datasets and exposure splits.}
We conduct all experiments on the EPIC-KITCHENS-100 Multi-Instance Retrieval
task (EK100-MIR)~\cite{damen2022rescaling}. We assess
EgoVLPv2~\cite{pramanick2023egovlpv2} pretrained on
EgoClip~\cite{lin2022egocentric}. Using the frozen Exposure Ledger
described in Section~\ref{sec:exposure-ledger}, the 9,668 evaluation queries
are partitioned into 8,648 Seen Composition (SC), 713 Unseen Composition (UC),
and 307 Unseen Atom (UA) queries.

\paragraph{Model and evaluation.}
The pretrained video and text encoders are kept frozen, and retrieval is performed
using their dual-encoder video--text similarity scores. We evaluate both
video-to-text (V$\rightarrow$T) and text-to-video (T$\rightarrow$V) retrieval
using mean Average Precision (mAP) and normalised Discounted Cumulative Gain
(nDCG). For domain adaptation, we train
only the residual adapters using the EK100 soft relevance objective.
We split the SC training data by video into 90\% subtraining and 10\%
development. For exposure reweighting, we fix the trained adapters and
bias network and vary only the exposure input at inference. We compare
the Original, Zeroed, and Shuffled conditions in
Section~\ref{sec:exposure-manipulation}.

\paragraph{Binding probes and statistics.}
We evaluate verb--noun binding using the noun-swap and verb-swap probes
defined in Section~\ref{sec:exposure-ledger}. The large binding probe contains
8,592 eligible SC queries and is used for the main continuous
margin--exposure analysis; the controlled small probe contains 200 queries.
We report binding accuracy and similarity margins together with Pearson and
Spearman correlations with log-scaled pair exposure. For SC/UC/UA retrieval
comparisons, we report group means with 95\% bootstrap confidence intervals.

\subsection{Retrieval Performance Scales with Pretraining Exposure}
\label{sec:retrieval-exposure}

We first examine whether zero-shot retrieval performance depends on whether
a verb--noun composition was observed during pretraining.

\begin{figure*}[t]
    \centering
    \includegraphics[width=1\textwidth]{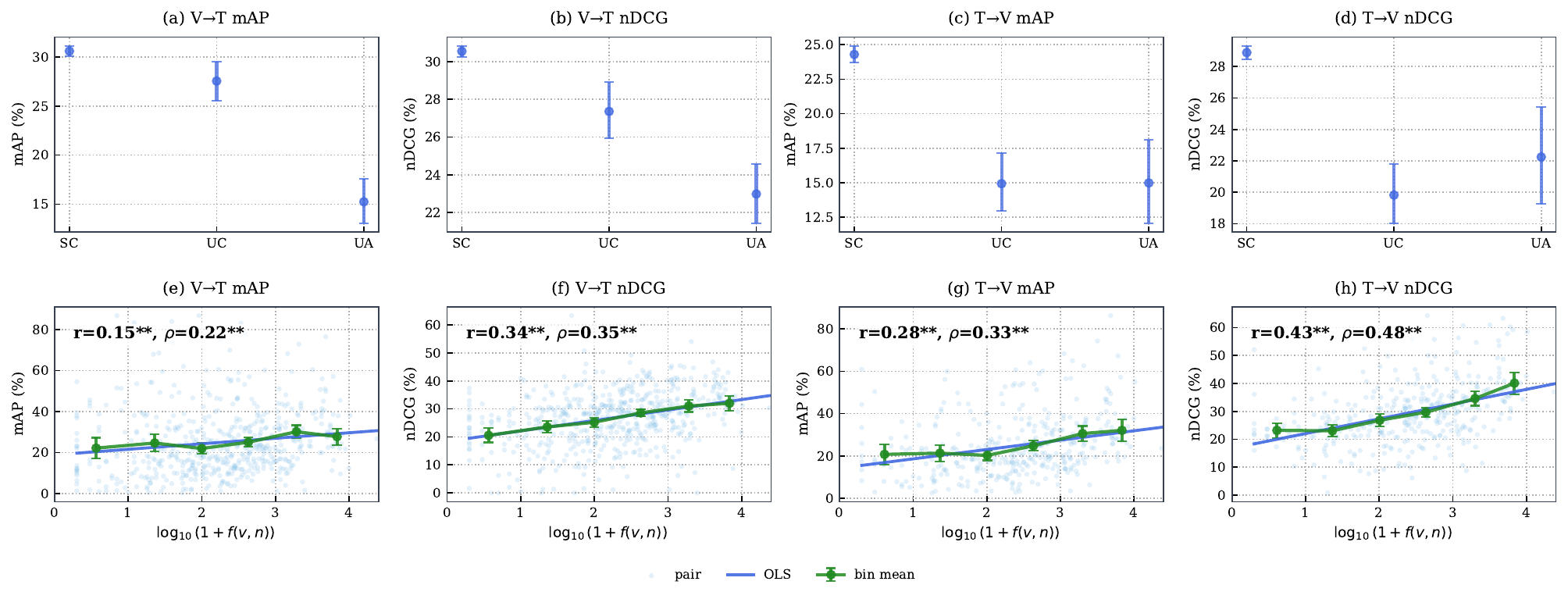}
    \caption{
    \textbf{Retrieval scales with pretraining composition exposure.}
    Top: zero-shot MIR across SC, UC, and UA. Bottom: pair-level retrieval
    versus log-scaled pair exposure within seen compositions. Markers show bin
    means with 95\% confidence intervals; lines show OLS fits. $\ast\ast$
    indicates $p<0.01$.
    }
    \label{fig:retrieval-exposure}
\end{figure*}

Zero-shot retrieval performance decreases when compositions are unseen. As
shown in Figure~\ref{fig:retrieval-exposure}(a--d), the model performs worse on
unseen compositions than on seen compositions in both retrieval directions.
For V$\rightarrow$T, both UC and UA underperform SC. For T$\rightarrow$V,
the drop is especially clear: mAP decreases from 24.30 on SC to 14.92 on UC,
which is already close to the 14.98 obtained on UA. These results show a substantial performance gap between compositions that
were seen and unseen during pretraining.

We next test whether retrieval also varies with the amount of exposure within
seen compositions. For each seen verb--noun composition, we regress pair-level
retrieval performance against $\log_{10}(1 + f(v,n))$.
Figure~\ref{fig:retrieval-exposure}(e--h) presents exposure-bin averages for
visualization, while Pearson and Spearman correlations are computed over the
original pair-level observations. Across all four direction--metric
combinations, retrieval increases with log-scaled pair exposure. The strongest
relationship appears for T$\rightarrow$V nDCG, with Pearson $r=0.43$ and
Spearman $\rho=0.48$.

\begin{table*}[t]
\centering
\small
\setlength{\tabcolsep}{7pt}
\renewcommand{\arraystretch}{1.12}
\begin{tabular}{llccccc}
\toprule
Direction
& Metric
& Atom-only $R^2$
& Atom+pair $R^2$
& $\Delta R^2$
& $\beta_{\mathrm{pair}}$
& HC3 $p$
\\
\midrule
V$\rightarrow$T
& mAP
& 0.012
& 0.025
& +0.013
& 3.117
& $0.033^{*}$
\\
V$\rightarrow$T
& nDCG
& 0.129
& 0.159
& +0.030
& 2.931
& $2.27\times10^{-4}$\textsuperscript{***}
\\
T$\rightarrow$V
& mAP
& 0.018
& 0.094
& +0.076
& 7.042
& $3.37\times10^{-5}$\textsuperscript{***}
\\
T$\rightarrow$V
& nDCG
& 0.073
& 0.203
& +0.131
& 7.056
& $3.18\times10^{-9}$\textsuperscript{***}
\\
\bottomrule
\end{tabular}
\vspace{4pt}
\caption{
\textbf{Composition exposure predicts retrieval beyond atom frequencies.}
The atom-only model uses verb and noun exposure, whereas
the atom+pair model additionally includes verb--noun pair
exposure. $\Delta R^2$ measures the additional explained
variance, and $\beta_{\mathrm{pair}}$ is the unstandardized
pair-exposure coefficient. HC3 $p$-values are
heteroskedasticity-robust.
$^{*}p<0.05$, $^{***}p<0.001$.
}
\label{tab:pair-regression}
\end{table*}

Finally, we examine whether pair frequency is associated with
retrieval performance after accounting for verb and noun frequencies.
Within seen compositions, we compare an atom-only model with an
atom+pair model that additionally includes $\log_{10}(1+f(v,n))$.
As shown in Table~\ref{tab:pair-regression}, the pair frequency
coefficient is positive in all four settings. The increase in
pair-level $R^2$ ranges from 0.013 for V$\rightarrow$T mAP to 0.131
for T$\rightarrow$V nDCG.

\subsection{Pair Exposure and Verb--Noun Binding Diagnostics}
\label{sec:binding-exposure}

We next ask whether exposure-related retrieval gains correspond to stronger
verb--noun binding. We use the controlled noun-swap and verb-swap probes
defined in Section~\ref{sec:exposure-ledger} on the frozen model. Accuracy
measures whether the correct caption is ranked above the corresponding foil,
whereas margin measures the strength of this preference.

For the main exposure-dependence analysis, we use the large val-SC subset
containing 8,592 seen-composition probes.

\begin{figure*}[t]
    \centering
    \includegraphics[width=0.88\textwidth]{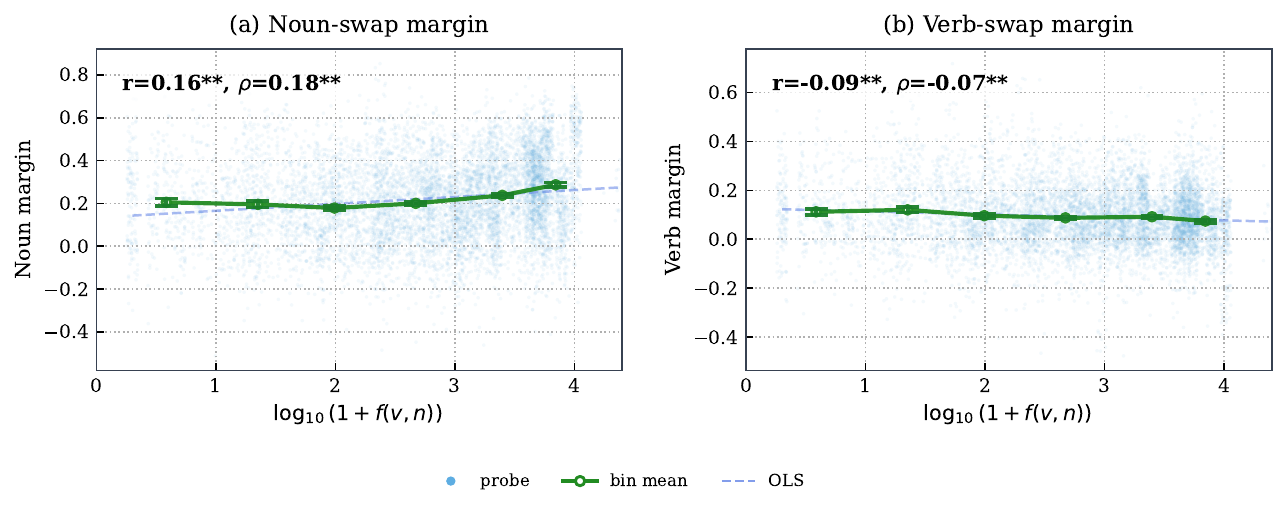}
    \caption{
    \textbf{Swap margins relate differently to pair exposure.}
    Within Seen Compositions, noun-swap (a) and verb-swap (b) margins are
    plotted against $\log_{10}(1+f(v,n))$ over 8,592 frozen-model probes.
    Light-blue points are individual probes, green circles are bin means with
    95\% confidence intervals, and dashed lines are OLS fits. Correlations are
    computed on the raw probes; $\ast\ast$ indicates $p<0.01$.
    }
    \label{fig:binding-exposure}
\end{figure*}

The frozen model shows a clear noun-dominant binding pattern. Although both
noun-swap and verb-swap accuracies are above chance, the margins reveal a much
stronger noun channel: the noun margin is 0.222, compared with 0.092 for the
verb margin. Thus, the noun channel has substantially stronger score
separation than the verb channel.

Figure~\ref{fig:binding-exposure} further shows that this binding structure is
exposure-asymmetric. Noun margin is positively associated with pair exposure
($\rho=0.181$, 95\% CI [0.160, 0.202]), whereas the verb-swap margin shows a weak
negative association ($\rho=-0.072$, 95\% CI [-0.093, -0.050]). Thus, higher pair exposure is associated with larger
noun-swap margins and slightly smaller verb-swap
margins. The exposure-related retrieval gain observed in
Section~\ref{sec:retrieval-exposure} is therefore not accompanied by
uniformly larger noun- and verb-swap margins.

\subsection{Retrieval-Oriented Interventions Do Not Restore Frozen Binding}
\label{sec:retrieval-interventions}

This section reports a seed-2026 diagnostic
intervention chain to keep the domain-adapted, verb-hard-negative, and
decorrelated adapters directly comparable.

We first test whether standard domain adaptation can restore binding. As shown in Table~\ref{tab:retrieval-interventions}, the domain-adapted adapter substantially improves MIR retrieval, with average mAP
(the mean of V$\rightarrow$T and T$\rightarrow$V) increasing from 26.67 for
the frozen model to 31.99. However, this retrieval improvement is accompanied
by a large decrease in the original noun-binding signal: noun margin drops from
0.222 to 0.054, and noun exposure correlation drops from $\rho=0.181$ to
$\rho=0.011$. Thus, improving MIR retrieval does not preserve the binding
signal present in the frozen model.

We then respond to the weak verb channel with verb-focused hard negatives,
replacing the verb while keeping the same noun. We select $\lambda=0.50$ based on a small ablation over verb hard-negative weights, as it provides the clearest local improvement in the verb channel. With $\lambda=0.50$, continued
verb-focused training produces a moderate local improvement: average mAP
increases from 31.99 to 33.33, and verb margin increases from 0.078 to 0.091.
However, this targeted improvement does not restore the full binding profile.
The noun margin reaches only 0.079, far below the frozen-model value of 0.222,
and verb exposure dependence becomes more negative ($\rho=-0.108$).
Verb-focused hard negatives therefore improve the targeted verb-discrimination
channel, but not full verb--noun binding.
Finally, we evaluate anchor-based noun-margin protection with and
without exposure decorrelation. In Table~\ref{tab:retrieval-interventions},
average mAP decreases from 32.22 at $\lambda_d=0$ to 31.71 at
$\lambda_d=100$. The corresponding noun-swap margins are 0.055 and
0.054, while the verb-swap margin is 0.078 in both settings. Both
noun-swap margins remain well below the frozen baseline of 0.222.

\begin{table*}[t]
\centering
\small
\setlength{\tabcolsep}{4.2pt}
\renewcommand{\arraystretch}{1.12}
\begin{tabular}{lcccccccc}
\toprule
&
\multicolumn{2}{c}{mAP}
&
\multicolumn{2}{c}{nDCG}
&
\multicolumn{2}{c}{Binding margin}
&
\multicolumn{2}{c}{Exposure corr.}
\\
\cmidrule(lr){2-3}
\cmidrule(lr){4-5}
\cmidrule(lr){6-7}
\cmidrule(lr){8-9}
Method
& V$\rightarrow$T
& T$\rightarrow$V
& V$\rightarrow$T
& T$\rightarrow$V
& Noun
& Verb
& $\rho_n$
& $\rho_v$
\\
\midrule
Frozen zero-shot
& 29.88
& 23.45
& 30.07
& 28.12
& \textbf{0.222}
& \textbf{0.092}
& 0.181
& $-0.072$
\\
\midrule
Domain adaptation
& 35.45
& 28.54
& 48.31
& 43.97
& 0.054
& 0.078
& 0.011
& $-0.069$
\\
\midrule
+ Verb hard negatives
& \textbf{36.35}
& \textbf{30.30}
& 46.98
& 43.38
& 0.079
& 0.091
& 0.028
& $-0.108$
\\
\midrule
+ Anchor + noun protect
& 35.89
& 28.55
& \textbf{48.43}
& 43.93
& 0.055
& 0.078
& 0.009
& $-0.071$
\\
+ Exposure decorr. $\lambda_d=50$
& 34.79
& 28.33
& 46.98
& 43.79
& 0.051
& 0.072
& 0.018
& $\textbf{-0.064}$
\\
+ Exposure decorr. $\lambda_d=100$
& 34.85
& 28.57
& 48.02
& \textbf{44.00}
& 0.054
& 0.078
& \textbf{0.008}
& $-0.071$
\\
\bottomrule
\end{tabular}
\vspace{5pt} 
\caption{
\textbf{Retrieval-oriented interventions improve retrieval without consistently
restoring verb--noun binding.}
We report V$\rightarrow$T and T$\rightarrow$V retrieval performance, noun and
verb binding margins, and their correlations with pair exposure.
}
\label{tab:retrieval-interventions}
\end{table*}

\subsection{Exposure Reweighting Changes Retrieval Without Binding Repair}
\label{sec:exposure-reweighting}

For the exposure score-bias counterfactuals, we perform the scoring-stage
intervention defined in Section~\ref{sec:exposure-manipulation} while keeping trained adapters and bias network fixed. Because the exposure bias is added
on the text side, it changes V$\rightarrow$T rankings, whereas
T$\rightarrow$V rankings remain unchanged by construction.

We compare three conditions: Original uses the true exposure vector, Zeroed
removes exposure, and Shuffled preserves the exposure distribution but breaks
the sample--exposure pairing. As shown in
Table~\ref{tab:exposure-reweighting}, using the true exposure vector increases V$\rightarrow$T mAP from 32.91 to 33.70 and nDCG from 43.04 to 44.22.
Within the exposure-relative splits, SC mAP increases from 33.94 to 34.97,
whereas UC mAP decreases from 26.14 to 24.51. Thus, exposure reweighting
primarily benefits compositions already observed during pretraining.

The Shuffled condition further shows that this gain depends on the correct
sample--exposure pairing. Compared with Original exposure, Shuffled exposure
reduces V$\rightarrow$T mAP from 33.70 to 32.56, nDCG from 44.22 to
42.80, and SC mAP from 34.97 to 33.56. In contrast, noun and verb binding
margins remain essentially unchanged across Original, Zeroed, and Shuffled
conditions. These results show that the correct sample-level exposure signal
can alter retrieval ranking without a corresponding improvement in
verb--noun binding. This interpretation is limited to counterfactual exposure replacement at a fixed trained checkpoint.

\begin{table*}[t]
\centering
\small
\setlength{\tabcolsep}{7pt}
\renewcommand{\arraystretch}{1.10}

\begin{tabular}{@{}lcccccc@{}}
\toprule
&
\multicolumn{2}{c}{V$\rightarrow$T Retrieval}
&
\multicolumn{2}{c}{V$\rightarrow$T split mAP}
&
\multicolumn{2}{c}{Binding Margin}
\\
\cmidrule(lr){2-3}
\cmidrule(lr){4-5}
\cmidrule(lr){6-7}
Exposure
& mAP
& nDCG
& SC
& UC
& Noun
& Verb
\\
\midrule

Zeroed
& 32.91
& 43.04
& 33.94
& 26.14
& 0.058
& 0.065
\\

\textbf{Original}
& \textbf{33.70}
& \textbf{44.22}
& \textbf{34.97}
& 24.51
& 0.058
& 0.064
\\

Shuffled
& 32.56
& 42.80
& 33.56
& 26.15
& 0.058
& 0.064
\\

\bottomrule
\end{tabular}
\vspace{5pt} 
\caption{
\textbf{Exposure reweighting changes V$\rightarrow$T retrieval without binding repair.}
At the same trained checkpoint, Original uses the true exposure vector,
Zeroed removes the exposure signal, and Shuffled breaks the sample--exposure
correspondence while preserving the exposure distribution.
T$\rightarrow$V is unchanged by construction and is omitted for compactness.
Binding margins remain essentially unchanged across conditions.
}
\label{tab:exposure-reweighting}
\end{table*}
\section{Conclusion}
\label{sec:conclusion}

In this work, we propose a new diagnostic protocol, Ego-ExBind, which
analyzes pretraining exposure and verb--noun compositional binding in
egocentric video--text retrieval. We demonstrate that zero-shot retrieval
performance is associated with pretraining composition exposure, with
retrieval performance increasing with log-scaled pair exposure beyond the
marginal frequencies of verbs and nouns. We further show that this exposure
dependence is asymmetric across binding channels: higher pair exposure is
associated with larger noun margins, but not larger verb margins.

We introduce a scoring stage exposure intervention to quantify its effects on retrieval performance and binding. We find that it improves overall video-to-text retrieval, with little change in the noun-swap and verb-swap margins. We also compare the Original, Zeroed, and Shuffled exposure conditions, showing that the retrieval gain depends on the correct sample level exposure signal.

Our findings reveal that gains in model retrieval performance do not
necessarily represent improved compositional understanding. Future VLM evaluations should
distinguish exposure familiarity from systematic compositional generalization.

%------------------------------------------------------------------------

\bibliography{egbib}
\end{document}